\def\year{2021}\relax
\documentclass[letterpaper]{article} 
\usepackage{aaai21}  
\usepackage{times}  
\usepackage{helvet} 
\usepackage{courier}  
\usepackage[hyphens]{url}  
\usepackage{graphicx} 
\usepackage{natbib}  
\usepackage{caption} 
\usepackage{algorithmicx, algpseudocode}
\usepackage{multirow}
\usepackage{algorithm}
\usepackage{booktabs}
\usepackage{color}
\usepackage{xcolor}
\usepackage{subcaption}
\usepackage{colortbl}

\title{Revisiting Iterative Back-Translation from the Perspective of Compositional Generalization}
\author{Yinuo Guo\textsuperscript{\rm 1}{\thanks{Work done during an internship at Microsoft Research Asia. The first two authors contributed equally.}}~, Hualei Zhu\textsuperscript{\rm 2}$^*$, Zeqi Lin\textsuperscript{\rm 3},  Bei Chen\textsuperscript{\rm 3}, Jian-Guang Lou\textsuperscript{\rm 3}, Dongmei Zhang\textsuperscript{\rm 3}\\}
\affiliations{
    \textsuperscript{\rm 1}Key Laboratory of Computational Linguistics, School of EECS, Peking University,\\
    \textsuperscript{\rm 2}School of Computer Science and Engineering, Beihang University\\
    \textsuperscript{\rm 3}Microsoft Research Asia \\
    \textsuperscript{\rm 1}gyn0806@pku.edu.cn, \textsuperscript{\rm 2}zhuhualei@buaa.edu.cn\\
    \textsuperscript{\rm 3}\{Zeqi.Lin, jlou, dongmeiz\}@microsoft.com
}
\begin{document}

\maketitle

\begin{abstract}
Human intelligence exhibits compositional generalization (i.e., the capacity to
understand and produce unseen combinations of seen components), but current neural seq2seq models lack such ability.
In this paper, we revisit \emph{iterative back-translation}, a simple yet effective semi-supervised method, to investigate whether and how it can improve compositional generalization.
In this work:
(1) We first empirically show that iterative back-translation substantially improves the performance on compositional generalization benchmarks (CFQ and SCAN).
(2) To understand why iterative back-translation is useful, we carefully examine the performance gains and find that iterative back-translation can increasingly correct errors in pseudo-parallel data.
(3) To further encourage this mechanism, we propose \emph{curriculum iterative back-translation}, which better improves the quality of pseudo-parallel data, thus further improving the performance.
\end{abstract}

\section{Introduction}
Human intelligence exhibits \emph{compositional generalization}, the ability to understand and produce unseen combinations of seen components \cite{chomsky1965aspects,fodor1988connectionism}.
For example, if a person knows the meaning of ``run'' and ``run twice'', once she learns the meaning of a new word ``jump'', she can immediately understand the meaning of ``jump twice''.
This compositional generalization is essential in human cognition, enabling humans to learn semantics from very limited data and extend to unseen data \cite{montague1974formal,partee1984compositionality,lake2017building}.
Therefore, it is widely believed that compositional generalization is a key property towards human-like AI \cite{mikolov2016roadmap,bengio2019towards,mollica2020composition}.
However, in recent years, accumulating evidences have shown that neural sequence-to-sequence (seq2seq) models exhibit limited compositional generalization ability~\cite{lake2018generalization,hupkes2020compositionality,keysers2020measuring,furrer2020compositional}.



On the other hand, semi-supervised learning \cite{chapelle2009semi} is a machine learning paradigm that alleviates the limited-data problem through exploiting a large amount of unlabelled data, which has been successfully used in many different areas such as machine translation \cite{sennrich2016improving,he2016dual,skorokhodov2018semi} and semantic parsing \cite{yin2018structvae,cao2019semantic}.
In sequence-to-sequence (seq2seq) tasks, unlabelled data (i.e., monolingual data) are usually cheap and abundant, containing a large amount of combinations that are unseen in limited labelled data (i.e., parallel data).
Therefore, we propose a hypothesis that \emph{semi-supervised learning can enable seq2seq models understand and produce much more combinations beyond labelled data, thus tackling the bottleneck of lacking compositional generalization}.
If this hypothesis holds true, the lack of compositional generalization would no longer be a bottleneck of seq2seq models, as we can simply tackle it through exploiting large-scale monolingual data with diverse combinations.

In this work, we focus on \emph{Iterative Back-Translation (IBT)} \cite{hoang2018iterative}, a simple yet effective semi-supervised method that has been successfully applied in machine translation.
The key idea behind it is to iteratively augment original parallel training data with pseudo-parallel data generated from monolingual data.
To our best knowledge, iterative back-translation has not been studied extensively from the perspective of compositional generalization.
This is partially because a concern about \emph{the quality of pseudo-parallel data}:
due to the problem of lacking compositional generalization, for non-parallel data with unseen combinations beyond the parallel training data, pseudo-parallel data generated from them will be error-prone.
It is natural to speculate that errors in pseudo-parallel data are going to be reinforced and then even harm the performance.

This paper broadens the understanding of iterative back-translation from the perspective of compositional generalization, through answering three research questions:
\textbf{RQ1. } How does iterative back-translation affect neural seq2seq models' ability to generalize to more combinations beyond parallel data?
\textbf{RQ2. } If iterative back-translation is useful from the perspective of compositional generalization, what is the key that contributes to its success?
\textbf{RQ3. } Is there a way to further improve the quality of pseudo-parallel data, thereby further improving the performance?

\textbf{Main Contributions.}
(1) We empirically show that iterative back-translation substantially improves the performance on compositional generalization benchmarks (CFQ and SCAN) (Section \ref{section:c1}).
(2) To understand what contributes to its success, we carefully examine the performance gains and observe that iterative back-translation is effective to correct errors in pseudo-parallel data  (Section \ref{section:c2}).
(3) Motivated by this analysis, we propose \emph{curriculum iterative back-translation} to further improve the quality of pseudo-parallel data, thereby improving the performance of iterative back-translation (Section \ref{section:c3}).

\section{Background}

\subsection{Compositional Generalization Benchmarks}
\label{section:tasks}

In recent years, the compositional generalization ability of DNN models has become a hot research problem in artificial intelligence, especially in Natural Language Understanding (NLU), because compositional generalization ability has been recognized as a basic but essential capability of human intelligence~\cite{lake2017building}.
Two benchmarks, SCAN \cite{lake2018generalization} and CFQ \cite{keysers2020measuring}, have been proposed for measuring the compositional generalization ability of different machine learning-based NLU systems.
Our experiments are also conducted on these two benchmarks.

\subsubsection{SCAN}
The SCAN dataset consists of input natural language commands (e.g., ``\emph{jump and look left twice}'') paired with output action sequences (e.g., ``\emph{JUMP LTURN LOOK LTURN LOOK}'').
Difficult tasks are proposed based on different data split settings, e.g.,
(1)~\textbf{ADD\_JUMP}: the pairs of train and test are split in terms of the primitive \emph{JUMP}.
The train set consists of \emph{(jump, JUMP)} and all pairs without the primitive \emph{JUMP}.
The rest forms the test set.
(2)~\textbf{LENGTH}: Pairs are split by action sequence length into train set ($\leq 22$ tokens) and test set ($\geq 24$ tokens). (3)~\textbf{AROUND\_RIGHT}: The phrase ``\emph{around right}" is held out from the train set, while ``around" and ``right" appear separately.
(4)~\textbf{OPPOSITE\_RIGHT}: This task is similar to AROUND\_RIGHT, while the held-out phrase is ``\emph{opposite right}''.

\subsubsection{CFQ}
The \emph{Complex Freebase Questions (CFQ)} benchmark \cite{keysers2020measuring} contains input natural language questions paired with their output meaning representations (SPARQL queries against the Freebase knowledge graph).
To comprehensively measure a learner's compositional generalization ability, CFQ dataset is splitted into train and test sets based on two principles:
(1) \emph{Minimizing primitive divergence}: all primitives present in the test set are also present in the training set, and the distribution of primitives in the training set is as similar as possible to their distribution in the test set.
(2) \emph{Maximizing compound divergence}: the distribution of compounds (i.e., logical substructures in SPARQL queries) in the training set is as different as possible from the distribution in the test set.
The second principle guarantees that the task is compositionally challenging.
Following these two principles, three different \emph{maximum compound divergence (MCD)} dataset splits are constructed.
The mean accuracy of standard neural seq2seq models on the MCD splits is below 20\%.
Moreover, this benchmark also constructs 3 MCD data splits for SCAN dataset, and the mean accuracy of standard neural seq2seq models is about 4\%.

\begin{figure}[t]
\centering
\includegraphics[width=0.8\columnwidth,clip=true]{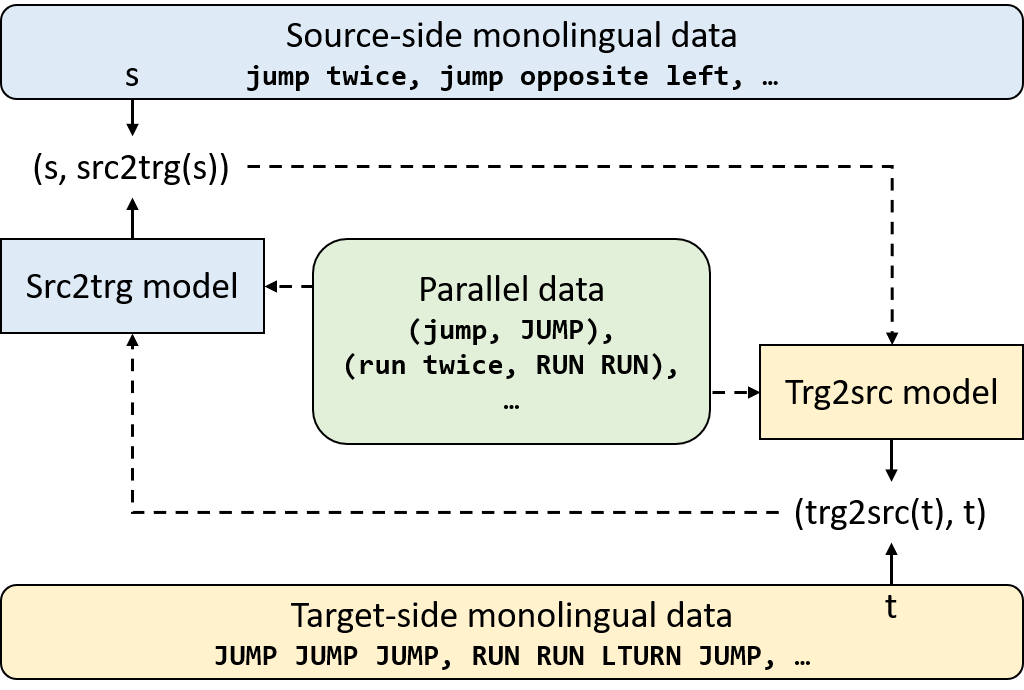}
\caption{Iterative back-translation: exploiting monolingual data to augment parallel training data.
The solid lines mean that the src2trg/trg2src model generates pseudo-parallel data from monolingual data;
the dashed lines mean that parallel and pseudo-parallel data are used to train the models.}
\label{fig:ibt}
\end{figure}

\subsection{Iterative Back-Translation~(IBT)}

Iterative back-translation~\cite{hoang2018iterative} has been proved as an effective method utilizing bi-directional monolingual data to improve the performance of machine translation models.
Algorithm~\ref{alg:ibt} describes the training process of iterative back-translation.
Firstly, the source-to-target (src2trg) model and target-to-source (trg2src) model are initially trained on the original parallel data ($D_p$) for $K$ steps (Line 1).
After this initial stage, each training step samples a batch of monolingual data from one direction and back-translate them to the other direction.
Then it would perform two mini-batches to train source-to-target model using the given parallel data and pseudo backtranslated parallel data separately, and also another two mini-batches for training target-to-source model.
Figure~\ref{fig:ibt} visualizes the iterative training process.

\begin{algorithm}[h]
\small
	\caption{Iterative Back-Translation}
	\textbf{Input:} parallel data $D_p$, monolingual data $D_{src}$, $D_{trg}$ \\
	\textbf{Output:} source-to-target model $M_{\rightarrow}$ and target-to-source model $M_{\leftarrow}$
	\begin{algorithmic}[1]
		\State \textbf{Initial Stage}: Separately train source-to-target model $M_{\rightarrow}$ and target-to-source model $M_{\leftarrow}$ on $D_{p}$ for K steps
		\While{$M_{\rightarrow}$ and $M_{\leftarrow}$ have not converged  \textbf{or} step $\leq$ N}
		    \State Sample a batch $B_{t}$ from $D_{trg}$ and use  $M_{\leftarrow}$ to create $B_p=\{(\hat{s}, t) | t\in B_t\}$ 
		    \State Sample a batch $B_{s}$ from $D_{src}$ and use  $M_{\rightarrow}$ to create $B_p'=\{(s, \hat{t}) | s\in B_s\}$
		    \State Train source-to-target model $M_{\rightarrow}$ on $D_{p}$
		    \State Train source-to-target model $M_{\rightarrow}$ on $B_p$
		    \State Train target-to-source model $M_{\leftarrow}$ on $D_{p}$
		    \State Train target-to-source model $M_{\leftarrow}$ on $B_{p}'$
	    \EndWhile
	\end{algorithmic}
\label{alg:ibt}
\end{algorithm}

\begin{table*}[htbp]
\small
\caption{Performance (accuracy) on SCAN Tasks. Cells with a white background are results obtained in previous papers; cells with a grey background are results obtained in this paper.}
\centering
\begin{tabular}{lccccccccc}
\toprule[1pt]
 Models  & ADD\_JUMP & LENGTH  &AROUND\_RIGHT& OPPOSITE\_RIGHT & MCD1 & MCD2 & MCD3\\
\midrule
LSTM+Att &$0.0\pm0.0$&$14.1$ & $0.0\pm0.0$ & $16.5\pm6.4$ &$6.5\pm3.0$ & $4.2\pm1.4$& $1.4\pm0.2$\\
Transformer &$1.0\pm0.6$& $0.0$& $53.3\pm10.9$ & $3.0\pm6.8$ & $0.4\pm0.2$ & $1.6\pm0.3$&$0.88\pm0.4$\\
Uni-Transformer& $0.3\pm0.3$ & $0.0$ & $47.0 \pm10.0$ &15.2$\pm$13.0& $0.5\pm0.1$ & $1.5\pm0.2$&$1.1\pm0.4$\\
\midrule
Syn-att&$91.0\pm27.4$&$15.2\pm0.7$&$28.9\pm34.8$&$10.5\pm8.8$&-&-&-\\
CGPS&$98.8\pm1.4$& $20.3\pm1.1$ &$83.2\pm13.2$ &$89.3\pm15.5$ &$1.2\pm1.0$ & $1.7\pm2.0$& $0.6\pm0.3$ \\
Equivariant&$99.1\pm0.0$&$15.9\pm3.2$&$\mathbf{92.0\pm0.2}$&-&-&-&-\\
GECA & $87.0$ &- & $82.0$&- &-&-&-\\
Meta-seq2seq&$99.9$&$99.9$&$16.6$&-&-&-&-\\
T5-11B&$98.3$&$3.3$&$49.2$&$\mathbf{99.1}$ &$7.9$&$2.4$ &$16.8$ \\
\midrule[1pt]
 GRU+Att~(Ours) & \cellcolor{lightgray}$0.6$ &\cellcolor{lightgray}$14.3$&\cellcolor{lightgray} $23.8$ & \cellcolor{lightgray}$0.27$  &\cellcolor{lightgray} $18.7$ &\cellcolor{lightgray}$32.0$& \cellcolor{lightgray}$42.4$ \\
\quad +mono30&\cellcolor{lightgray}$\mathbf{99.6}$&\cellcolor{lightgray}$\mathbf{77.7}$&\cellcolor{lightgray}$37.8$&\cellcolor{lightgray}$95.8$ &\cellcolor{lightgray} $\mathbf{64.3}$&\cellcolor{lightgray}$\mathbf{80.8}$&\cellcolor{lightgray}$\mathbf{52.2}$ \\
\midrule
\quad +mono100&\cellcolor{lightgray}$100.0$&\cellcolor{lightgray}$99.9$&\cellcolor{lightgray}$42.9$&\cellcolor{lightgray}$100.0$&\cellcolor{lightgray}$87.1$&\cellcolor{lightgray}$99.0$&\cellcolor{lightgray}$64.7$\\
\quad +transductive&\cellcolor{lightgray}$100.0$&\cellcolor{lightgray}$99.3$&\cellcolor{lightgray}$34.5$&\cellcolor{lightgray}$100.0$&\cellcolor{lightgray}$74.8$ &\cellcolor{lightgray}$99.7$ &\cellcolor{lightgray}$77.4$  \\
\bottomrule[1pt]
\end{tabular}
\label{tab:scan}
\end{table*}

\section{IBT for Compositional Generalization}
\label{section:c1}

To examine the effectiveness of IBT for compositional generalization, we conduct a series of experiments on two compositional generalization benchmarks (SCAN and CFQ, introduced in Section~\ref{section:tasks}).

\subsection{Setup}
\label{sec:ibt_setup}
\subsubsection{Monolingual Data Settings}
In our experiments, we conduct different monolingual data settings as follows:
\begin{itemize}
    \item \textbf{+transductive}:
    \emph{use all test data as monolingual data.}
    We use this setting to explore how iterative back-translation performs with the most ideal monolingual data.
    \item \textbf{+mono100}:
    \emph{use all dev data as monolingual data.}
    In this setting, test data do not appear in the monolingual data, while the monolingual data contains much more unseen combinations beyond the parallel training data.
    \item \textbf{+mono30}:
    \emph{randomly sample 30\% source sequences and 30\% target sequences from dev data.}
    This setting is more realistic than ``+tranductive'' and ``+mono100'', because there is no implicit correspondence between source-side and target-side monolingual data (i.e., for a sequence in source-side monolingual data, the target-side monolingual data do not necessarily contain its corresponding target sequence, and vice versa).
    Therefore, in our analysis of experimental results, we regard results of \textbf{``+mono30"} as the actual performance of iterative back-translation.
\end{itemize}

Note that because there are no dev data in SCAN tasks, we randomly hold out half of the original test data as the dev data (the held-out dev data and the remaining test data are not overlapped).

\subsubsection{Implementation Details} We implement iterative back-translation based on the code of UNdreaMT\footnote{https://github.com/artetxem/undreamt}~\cite{artetxe2018iclr}. For CFQ, we use a 2-layer GRU encoder-decoder model equipped with attention. We set the size of both word embedding and hidden states to 300. We use a dropout layer with the rate of 0.5 and the training process lasts 30000 iterations with batch size 128. For SCAN, we also use a 2-layer GRU encoder-decoder model with attention. Both of the embedding size and hidden size are set to 200. We use a dropout layer with the rate of 0.5 and the training process lasts 35000 iterations with batch size 64. We set $K=5000$ steps in Algorithm~\ref{alg:ibt} for both CFQ and SCAN benchmarks.
\subsubsection{Baselines}
(i) \textbf{General seq2seq models}:~LSTM, Transformer and Uni-Transformer~\cite{keysers2020measuring}. (ii)~\textbf{SOTA models on SCAN/CFQ benchmarks}:  Syn-att~\cite{russin2019compositional}, CGPS~\cite{li2019compositional},  Equivariant~\cite{gordon2019permutation}, GECA~\cite{andreas-2020-good}, Meta-seq2seq~\cite{lake2019compositional} and T5-11B~\cite{furrer2020compositional}. 

\begin{table}[tbp]
\small
\caption{Performance (accuracy) on CFQ tasks.}
\centering
\begin{tabular}{lcccc}
\toprule[1pt]
 Models & MCD1 & MCD2 & MCD3 \\
\midrule
LSTM+Attn & $28.9\pm1.8$ & $5.0\pm0.8$ & $10.8\pm0.6$  \\
Transformer & $34.9\pm1.1$ & $8.2\pm0.3$ & $10.6\pm1.1$ \\
Uni-Transformer & $37.4\pm2.2$ & $8.1\pm1.6$ & $11.3\pm0.3$\\
CGPS & $13.2\pm3.9$ & $1.6\pm0.8$ & $6.6\pm0.6$\\
T5-11B & $61.4\pm4.8$&$30.1\pm2.2$&$31.2\pm5.7$\\
\midrule[1pt]
GRU+Attn~(Ours) &\cellcolor{lightgray}$32.6\pm0.22$ &\cellcolor{lightgray}$6.0\pm0.25$ & \cellcolor{lightgray}$9.5\pm0.25$\\
\quad +mono30 & \cellcolor{lightgray}$\mathbf{64.8\pm4.4}$&\cellcolor{lightgray}$\mathbf{57.8\pm4.9}$&\cellcolor{lightgray}$\mathbf{64.6\pm4.9}$\\
\midrule
\quad +mono100&\cellcolor{lightgray}$83.2\pm3.1$&\cellcolor{lightgray}$71.5\pm6.9$&\cellcolor{lightgray}$81.3\pm1.6$\\
\quad +transductive &\cellcolor{lightgray}$88.4\pm0.7$ &\cellcolor{lightgray}$81.6\pm6.5$ &\cellcolor{lightgray}$88.2\pm2.2$\\
\bottomrule[1pt]
\end{tabular}
\label{tab:cfq}
\end{table}

\subsection{{Observations}}
\begin{figure*}[t]
\small
\centering
\includegraphics[width=0.9\textwidth]{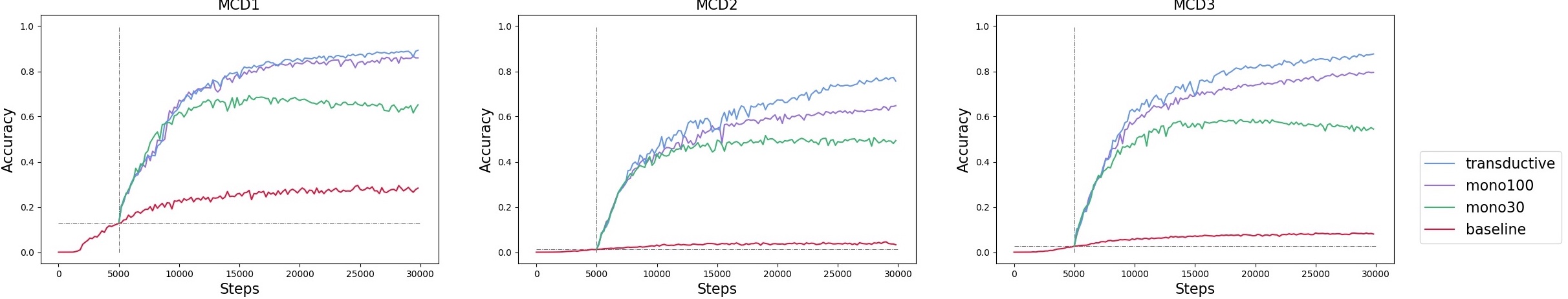}
\caption{Quality~(accuracy scores) of generated target data during the training procedure.}
\label{fig:accuracy_scores}
\end{figure*}
\begin{figure*}
\centering
\includegraphics[width=0.9\textwidth]{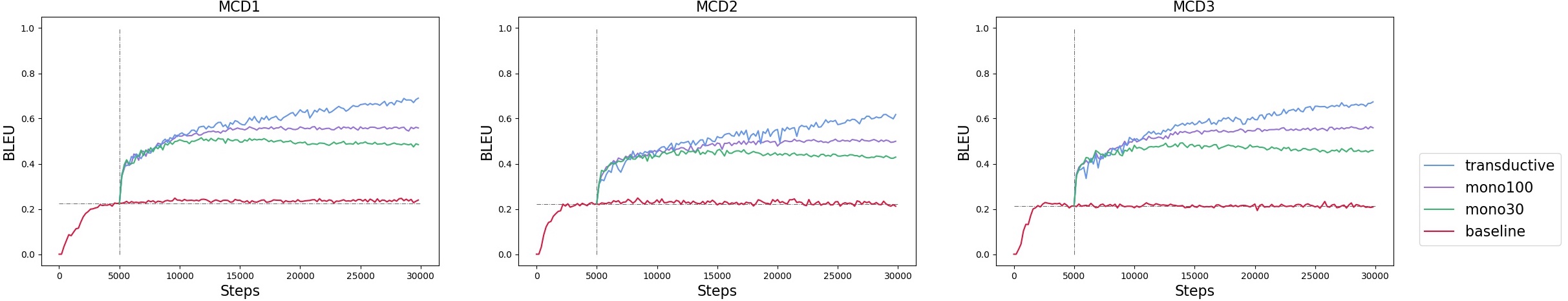}
\caption{Quality~(BLEU scores) of generated source data during the training procedure.}
\label{fig:bleu_scores}
\end{figure*}
We report the experimental results in Table~\ref{tab:scan} (SCAN) and~\ref{tab:cfq} (CFQ).
Notice that we pay more attention to MCD splits (on both SCAN and CFQ), since it has been proven that they are more comprehensive for evaluating compositional generalization than other splits \cite{keysers2020measuring,furrer2020compositional}.
Our observations are as follows:

\textit{1.Iterative Back-translation substantially improves the performance on compositional generalization benchmarks.}
Firstly, IBT can always bring large performance gains to our baseline (GRU+Att).
Then, compared to the previous state-of-the-art models, IBT can consistently achieve remarkable performance on all tasks, while those SOTA models only show impressive performances on ADD\_JUMP / LENGTH / AROUND\_RIGHT / OPPOSITE\_RIGHT tasks of SCAN, but perform poorly on MCD tasks of SCAN/CFQ.
Moreover, IBT also achieves higher performance than T5-11B, which is a strong baseline that incorporates rich external knowledge during pre-training stage.



\textit{2. Better monolingual data, better results.} 
As described in section~\ref{sec:ibt_setup}, the quality of monolingual data are gradually improved from  ``+mono30" to ``+mono100" then to ``+transductive". In most cases, it's as expected that ``+transductive" performs better than ``+mono100", and ``+mono100" performs better than ``+mono30".


\section{Secrets behind Iterative Back-Translation}
\label{section:c2}

Section \ref{section:c1} shows that iterative back-translation substantially improves seq2seq models' ability to generalize to more combinations beyond parallel data, but it is still unclear how and why iterative back-translation succeeds (RQ 2).

To answer this research question, we first empirically analyze pseudo-parallel data quality during the training process of iterative back-translation, and find that errors are increasingly corrected (Section \ref{section:quality}).
Then, we conduct ablation experiments to further illustrate this observation.
We find that:
\emph{even the error-prone pseudo-parallel data are beneficial.}
We speculate the reason is that knowledge of unseen combinations are implicitly injected into the model, thereby the src2trg model and the trg2src model can boost each other rather than harm each other (Section \ref{section:static}).
Moreover, pseudo-parallel data in iterative back-translation contain perturbations, helping models correct errors more effectively (Section \ref{section:otf}).
In this section, all experiments are conducted on the CFQ benchmark, as it is much more complex and comprehensive than SCAN.

\begin{figure*}[t]
\small
\centering	
    \begin{subfigure}{.35\textwidth}
    	\centering
    	\includegraphics[width=\textwidth]{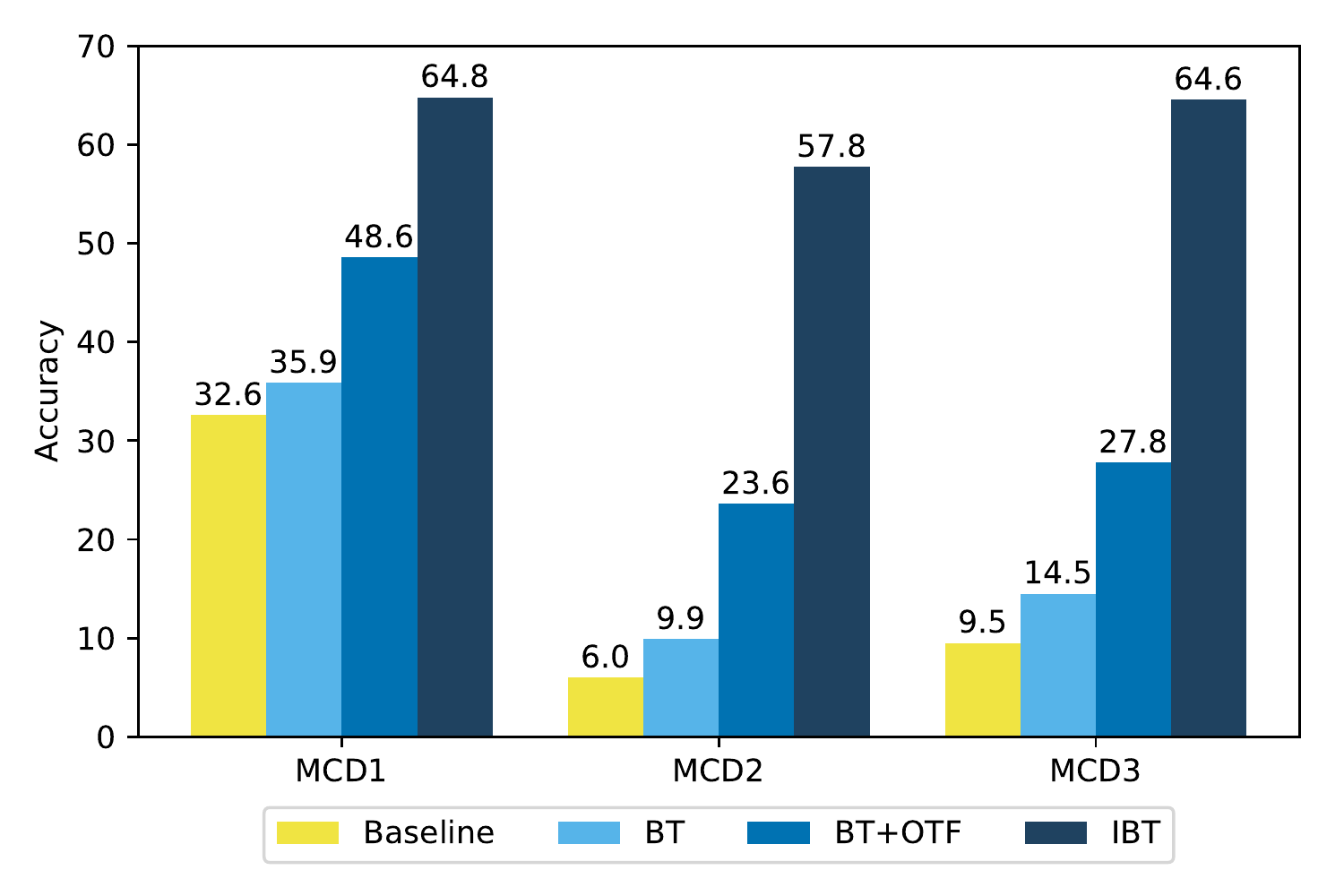}
    	\caption{Accuracy of Src2trg models}
    	\label{fig:acc}
    \end{subfigure}
    \begin{subfigure}{.35\textwidth}
    	\centering
    	\includegraphics[width=\textwidth]{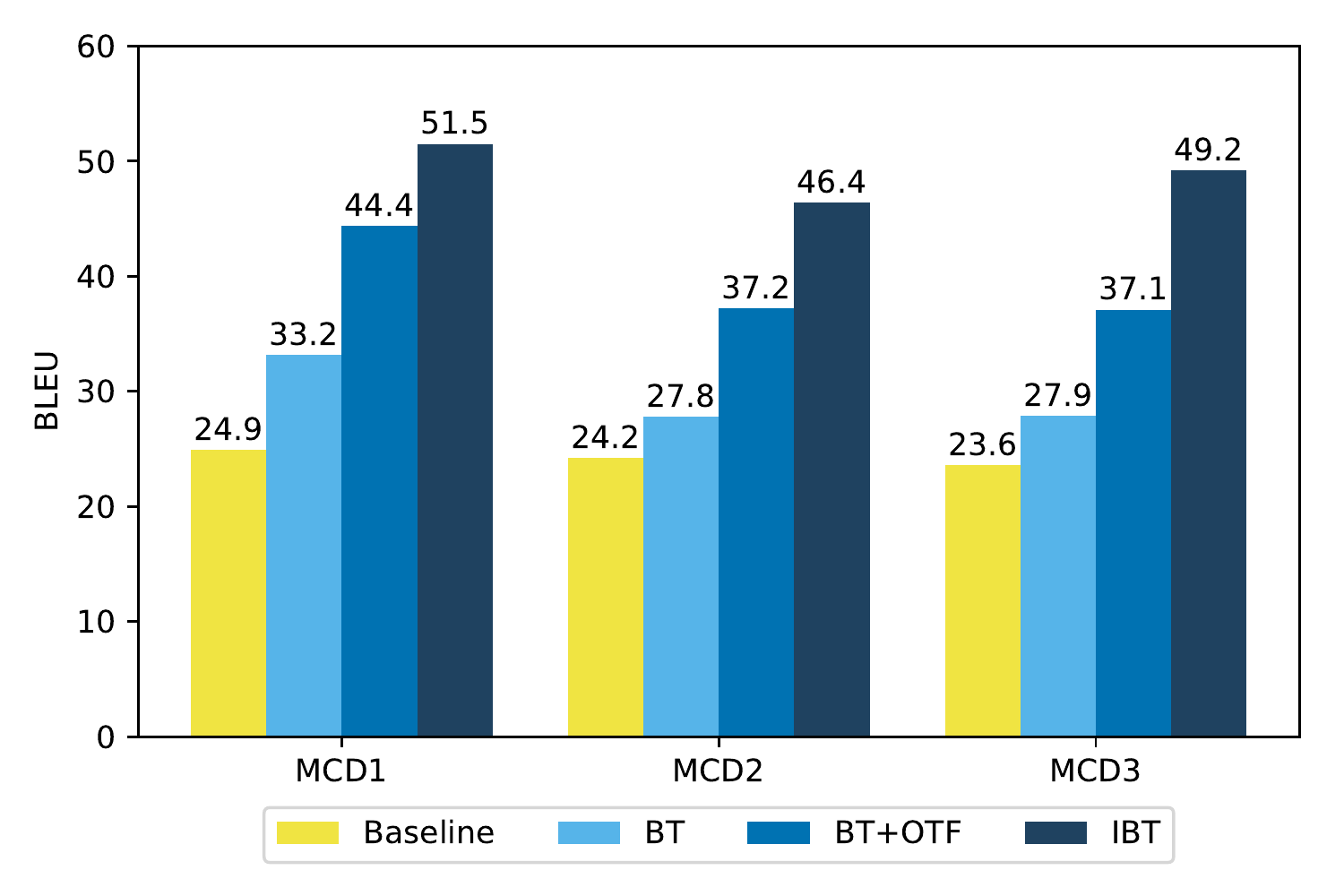}
    	\caption{BLEU of Trg2src models}
    	\label{fig:bleu}
    \end{subfigure}
\caption{Ablation experiments for understanding the key factors contribute to the performance gain.
The comparison of baseline and BT indicates that even error-prone pseudo-parallel data are beneficial due to the injected implicit knowledge of unseen combinations (Section \ref{section:static}).
The comparison of BT and BT+OTF indicates that perturbations brought by the on-the-fly mechanism can prevent models learning specific incorrect bias, thus improving the performance (Section \ref{section:otf}).
IBT performs best because it uses both source-side and target-side monolingual data while others use only target-side monolingual data.}
\label{fig:stanford_res}
\end{figure*}

\subsection{Quality of Pseudo-Parallel Data}
\label{section:quality}

We use Figure \ref{fig:accuracy_scores} and \ref{fig:bleu_scores} to get a better sense of the quality of pseudo-parallel data during the training process.
The x-axis represents the number of training steps: in the first 5000 steps, we train the src2trg model and the trg2src model with only parallel data;
the iterative back-translation process starts from step 5001 (dashed vertical lines).
For each step, we use the model to generate pseudo-parallel data from all monolingual data, and then evaluate the generated data quality.
Specifically:
the src2trg model generates a SPARQL query for each natural language utterance in the source-side monolingual data, and we define the quality of these generated SPARQL queries (i.e., target-side sequences) as the accuracy of them;
the trg2src model generates a natural language utterance for each SPARQL query in the target-side monolingual data, and we define the quality of these generated natural language utterances (i.e., source-side sequences) as the BLEU score~\cite{papineni2002bleu} of them.

We observe that \emph{iterative back-translation can increasingly correct errors in pseudo-parallel data.}
Even though the pseudo-parallel data are error-prone at the end of initial stage~(dashed horizontal lines), the accuracy/BLEU score has increased dramatically since the iterative back-translation process starts.




\subsection{Impact of Error-Prone Pseudo-Parallel Data}
\label{section:static}

To explore the reasons why iterative back-translation can increasingly correct error pseudo-parallel data, we investigate whether models can benefit from the initially generated pseudo-parallel data which are error-prone.
If yes, errors are going to be reduced during the iterative training process;
otherwise, errors are going to be reinforced.

Toward this end, we conduct two ablation experiments with ``+mono30" monolingual data setting:

\begin{itemize}
\item \textbf{BT-Src2trg}: the standard back-translation method for training the src2trg model (i.e., BT in Figure \ref{fig:acc}).
It consists of three phases:
first, train a src2trg model and a trg2src model using parallel data;
then, generate pseudo-parallel data from target-side monolingual data using the trg2src model;
finally, tune the src2trg model using the union of parallel data and pseudo-parallel data.
\item \textbf{BT-Trg2src}: the standard back-translation method for training the trg2src model (i.e., BT in Figure \ref{fig:bleu}).
It also consists of three phrases dual to those in BT-Src2trg.
\end{itemize}





According to the performance of BT and baseline (i.e., src2trg/trg2src models trained only on parallel data) in Figure \ref{fig:stanford_res}, we observe that \emph{even error-prone pseudo-parallel data can be beneficial.}
Specifically, though the initial pseudo-parallel data are error-prone (on average, 21.9\% BLEU for Trg2src and 16.0\% accuracy for Src2trg), they can still improve each other (5.4\% BLEU and 4.0\% accuracy gains for Trg2src and Src2trg models, respectively).

A reasonable illustration for the observation is that: even though pseudo-parallel data are error-prone, it can still implicitly inject the knowledge of unseen combinations into the src2trg/trg2src models, thereby boosting these models rather than harming them.

\subsection{Impact of Perturbations}
\label{section:otf}


Previous studies show that \emph{perturbations (or noises)} in pseudo data play an important role in semi-supervised learning \cite{zhang2016exploiting,edunov2018understanding,he2019revisiting,xie2020self}.

In iterative back-translation, the on-the-fly mechanism (i.e., models will be tuned each mini-batch on-the-fly with a batch of pseudo-parallel data) brings perturbations to pseudo-parallel data.
Different batches of pseudo-parallel data are produced by different models at different steps, rather than the same model.
This makes errors more diverse, thus preventing the src2trg/trg2src model learning specific incorrect bias from specific dual model.
Therefore, it is reasonable to speculate that some performance gains are brought by such perturbations.

We conduct two ablation experiments (with ``+mono30" monolingual data setting) to study the impact of such perturbations:

\begin{itemize}
\item \textbf{BT-Src2trg-with-OTF}:
this method can be seen as iterative back-translation without source-side monolingual data (i.e., BT+OTF in Figure \ref{fig:acc}).
Specifically, during the iterative training process, the trg2src model will only be tuned with parallel data.
As Figure \ref{fig:bleu_scores} shows, the quality of pseudo-parallel data produced by the trg2src model would keep error-prone if it is only tuned on parallel data (the baseline curves in Figure \ref{fig:bleu_scores}).
We want to see whether the src2trg model could be improved if the trg2src model keeps providing error-prone pseudo-parallel data.
\item \textbf{BT-Trg2src-with-OTF}:
this method can be seen as iterative back-translation without target-side monolingual data (i.e., BT+OTF in Figure \ref{fig:bleu}).
We want to see whether the trg2src model could be improved if the src2trg model keeps providing error-prone pseudo-parallel data.
\end{itemize}



According to the performance of BT and BT+OTF in Figure \ref{fig:stanford_res}, we find that perturbations brought by the on-the-fly mechanism do help models benefit from error-prone pseudo-parallel data more effectively.
For example, on average, the accuracy of BT-Src2trg-with-OTF is 33.3\%, while the accuracy of BT-Src2trg is only 20.1\%.
According to our statistics, each sequence in target-side monolingual data has 49.3 different back-translations during the iterative training process.
The perturbations bring 13.3\% performance gain, even if the quality of pseudo-parallel data has almost no improvement (see the baseline curves in Figure \ref{fig:bleu_scores}).

This experimental results support our hypothesis that \emph{perturbations brought by the on-the-fly mechanism contributes a lot to the success of iterative back-translation in compositional generalization benchmarks}.

\section{Curriculum Iterative Back-Translation}
\label{section:c3}

As discussed in Section \ref{section:c2}, the major reason for why iterative back-translation succeeds is that it can increasingly correct errors in pseudo-parallel data.
Based on this observation, it is reasonable to speculate that iterative back-translation can be further improved through injecting inductive bias that can help errors be reduced more efficiently.
Towards this end, we consider curriculum learning~\cite{bengio2009curriculum} as a potential solution: start out iterative back-translation with easy monolingual data (of which the generated pseudo-parallel data are less error-prone), then gradually increase the difficulty of monolingual data during the training process.
Based on this intuition, we propose \emph{Curriculum Iterative Back-Translation (CIBT)}.

\begin{table*}[t]
\small
    \centering
    \caption{Performance~(accuracy) of curriculum iterative back-translation.}
	\begin{tabular}{lcccccc}
	    \hline
    	\multirow{2}{*}{} & \multirow{2}{*}{IBT} & \multicolumn{5}{c}{CIBT with hyperparameter $c$ (steps in each stage)} \\ \cline{3-7}
    	 &  & 2000 & 2500 & 3000 & 3500 & 4000 \\ \cline{1-1} \cline{2-2} \cline{3-7}
         MCD1 & $  64.8 \pm 4.4 $ & $  66.1 \pm 5.0 $ & $  66.0 \pm 4.8 $ & $  \textbf{66.6} \pm 5.4 $ & $  65.9 \pm 3.7 $ & $  65.4 \pm 3.8 $ \\ 
        MCD2 & $  57.8 \pm 4.9 $ & $  68.6 \pm 2.6 $ & $  \textbf{69.1} \pm 3.1 $ & $  68.0 \pm 1.9 $ & $  66.8 \pm 2.4 $ & $  65.4 \pm 3.1 $ \\
        MCD3 & $  64.6 \pm 4.9 $ & $  70.2 \pm 4.9 $ & $  68.4 \pm 7.0 $ & $  \textbf{70.4} \pm 4.8 $ & $  69.2 \pm 4.1 $ & $  67.0 \pm 6.3 $ \\ 
        Mean & $  62.4 \pm 6.1 $ & $  \textbf{68.3} \pm 4.1 $ & $  67.8 \pm 4.7 $ & $  \textbf{68.3} \pm 4.1 $ & $  67.3 \pm 3.4 $ & $  65.9 \pm 4.1 $ \\ \hline
    \end{tabular}
	\label{cl_results_summary}
\end{table*}

\subsection{Method}

\subsubsection{Problem Formulation}

We denote source-side and target-side monolingual data in iterative back-translation as $D_{src}$ and $D_{trg}$, respectively.
Suppose that we have a simple heuristic algorithm $\mathcal{C}$, which can divide $D_{src}$ into $n$ parts (denoted as $D_{src}^{(1)}$, $D_{src}^{(2)}$, ..., $D_{src}^{(n)}$), and $D_{trg}$ can also be divided into $n$ parts (denoted as $D_{trg}^{(1)}$, $D_{trg}^{(2)}$, ..., $D_{trg}^{(n)}$).
They are sorted by difficulty in ascending order:
the initial src2trg model performs roughly better on $D_{src}^{(i)}$ than on $D_{src}^{(j)}$ if $i<j$;
and the initial trg2src model performs roughly better on $D_{trg}^{(i)}$ than on $D_{trg}^{(j)}$ if $i<j$.
We define that $(D_{src}^{(1)}, D_{trg}^{(1)}), (D_{src}^{(2)}, D_{trg}^{(2)}), ..., (D_{src}^{(n)}, D_{trg}^{(n)})$ are $n$ curriculums from simple to difficult.
Our goal is to improve the performance of iterative back-translation with this additional curriculum setting.

\subsubsection{Method Details}

Algorithm \ref{cl alg} details the workflow of curriculum iterative back-translation.

\begin{algorithm}[!h]
\small
	\caption{Curriculum Iterative Back-Translation}
	\label{cl alg}
	\textbf{Input:} $M_{\rightarrow}$ and $M_{\leftarrow}$, the src2trg and trg2src models that have been initially trained on parallel data; $D_{src}$ and $D_{trg}$, source-side and target-side monolingual data; $\mathcal{C}$, curriculum segmentation algorithm; $c$, the number of steps each stage should be trained; $n$, number of curriculums.
	\begin{algorithmic}[1]
		\State Use $\mathcal{C}$ to split $D_{src}$ into $D^{(1)}_{src}, D^{(2)}_{src}, ..., D^{(n)}_{src}$
		\State Use $\mathcal{C}$ to split $D_{trg}$ into $D^{(1)}_{trg}, D^{(2)}_{trg}, ..., D^{(n)}_{trg}$
		\State $D'_{src}=\emptyset, D'_{trg}=\emptyset$
		\For{$t= 1, 2, ..., n$} \algorithmiccomment{$n$ stages}
		\State $D'_{src} = D'_{src} \cup D^{(t)}_{src}$
		\State $D'_{trg} = D'_{trg} \cup D^{(t)}_{trg}$
		\State Iterative back-translation ($c$ steps) to update $M_{\rightarrow}$ and $M_{\leftarrow}$, with $D'_{src}$ and $D'_{trg}$ as monolingual data
		\EndFor
	\end{algorithmic}
\end{algorithm}

\subsection{Experimental Setup}
We evaluate the effectiveness of curriculum iterative back-translation on the CFQ ``+mono30" setting.

\subsubsection{Curriculum Segmentation.}
Defining the curriculum segmentation algorithm $\mathcal{C}$ is critical in our method.
In previous research work on curriculum learning, many metrics for curriculum segmentation have been proposed, such as length, frequency, and perplexity.

In curriculum iterative back-translation, $\mathcal{C}$ is used to split both $D_{src}$ and $D_{trg}$.
We denote that $\mathcal{C}_i(s)$ is an indicator that takes the value 1 if and only if $\mathcal{C}$ put the sample $s$ into the $i$-th curriculum.
It is a straightforward assumption that: for each $i = 1, 2, ..., n$, the performance of curriculum iterative back-translation would be better if $\mathcal{C}_i(x)=\mathcal{C}_i(y)$ for any source-side sequence $x$ paired with its ground truth target-side sequence $y$.
Intuitively, we want to ensure that $x$ and $y$ are in the same curriculum.

Based on this principle, we define $\mathcal{C}$ based on the number of entities occurring in each source/target sample.
For example, consider the source sequence \textit{``Were M1 and M3 distributed by M0's employer and distributed by M2?"}.
It contains 4 entities (M0, M1, M2, and M3), and its corresponding target sequence (a SPARQL query) also contains these 4 entities.
Therefore, we put them into the same curriculum.
Specifically, we define 6 curriculums, which contain monolingual data with $(\leq 1) / 2 / 3 / 4 / 5 / (\geq 6)$ entities.

\begin{figure}[t]
\centering
\small
\includegraphics[width=0.9\columnwidth,clip=true]{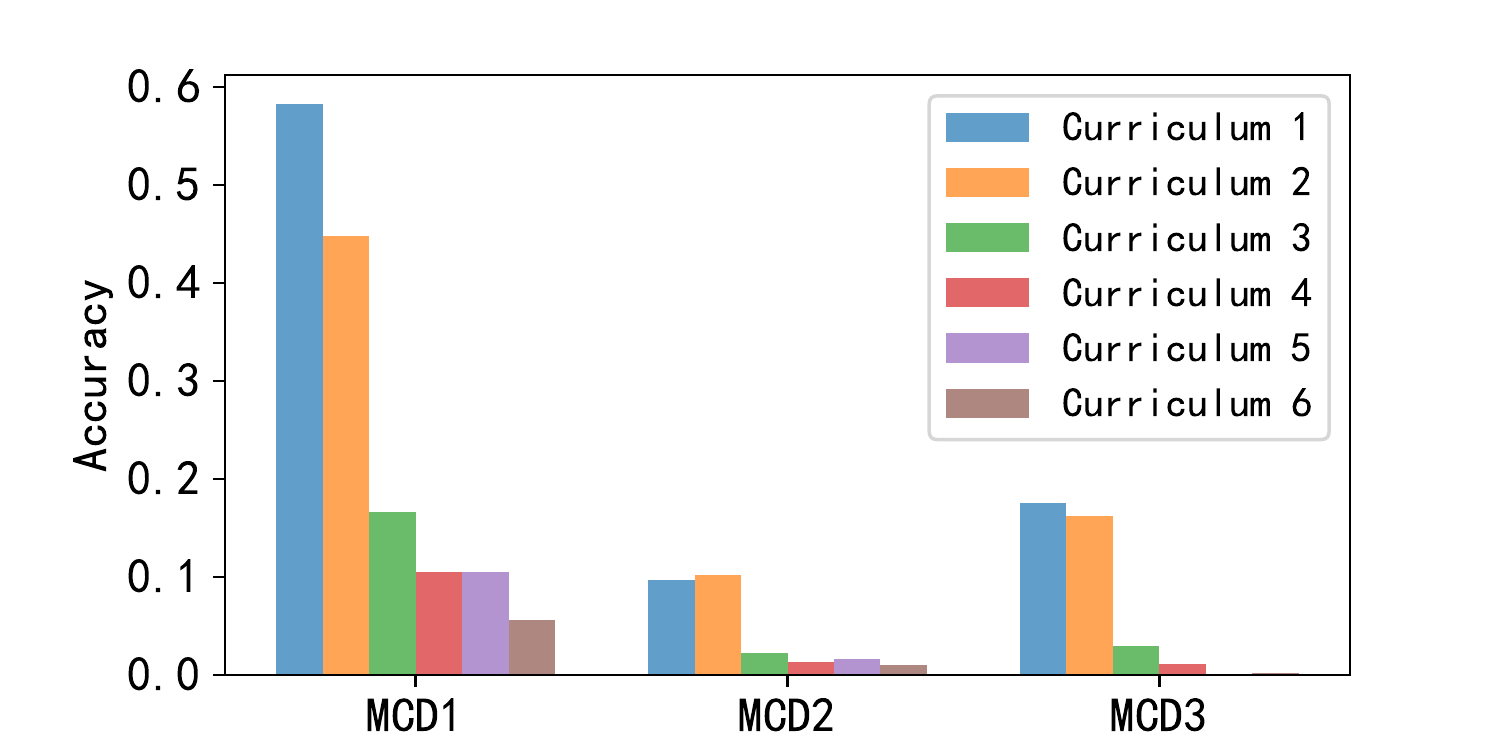} 
\caption{Performance of baseline model in each curriculum.}
\label{baseline_acc_in_each_curriculum}
\end{figure}

Figure \ref{baseline_acc_in_each_curriculum} shows that: in this entity-based curriculum setting, curriculum 1-6 are indeed arranged in order from simple to difficult.
Specifically, we evaluate the performance of the baseline model (i.e., the seq2seq model trained on parallel data) on $D_{src}^{(1)}, D_{src}^{(2)}, ..., D_{src}^{(6)}$, respectively.
We observe that the baseline model performs better on curriculum 1-2 than on curriculum 3-6, which indicates that this entity-based curriculum setting is reasonable.

\subsubsection{Hyperparameters}
We evaluate the performance of curriculum iterative back-translation with varying $c$ (the number of steps each stage should be trained): 2000/2500/3000/3500/4000.
To be fair, after $n\times c$ steps, we continue to train on complete monolingual data (i.e., $D_{src}$ and $D_{trg}$) for $(25,000-n\times c)$ steps, thus keeping the number of overall training steps the same.

\subsection{Results and Analysis}
\subsubsection{Curriculum learning benefits iterative back-translation.}

\begin{figure}[t]
\small
\centering
\includegraphics[width=0.7\columnwidth,clip=true]{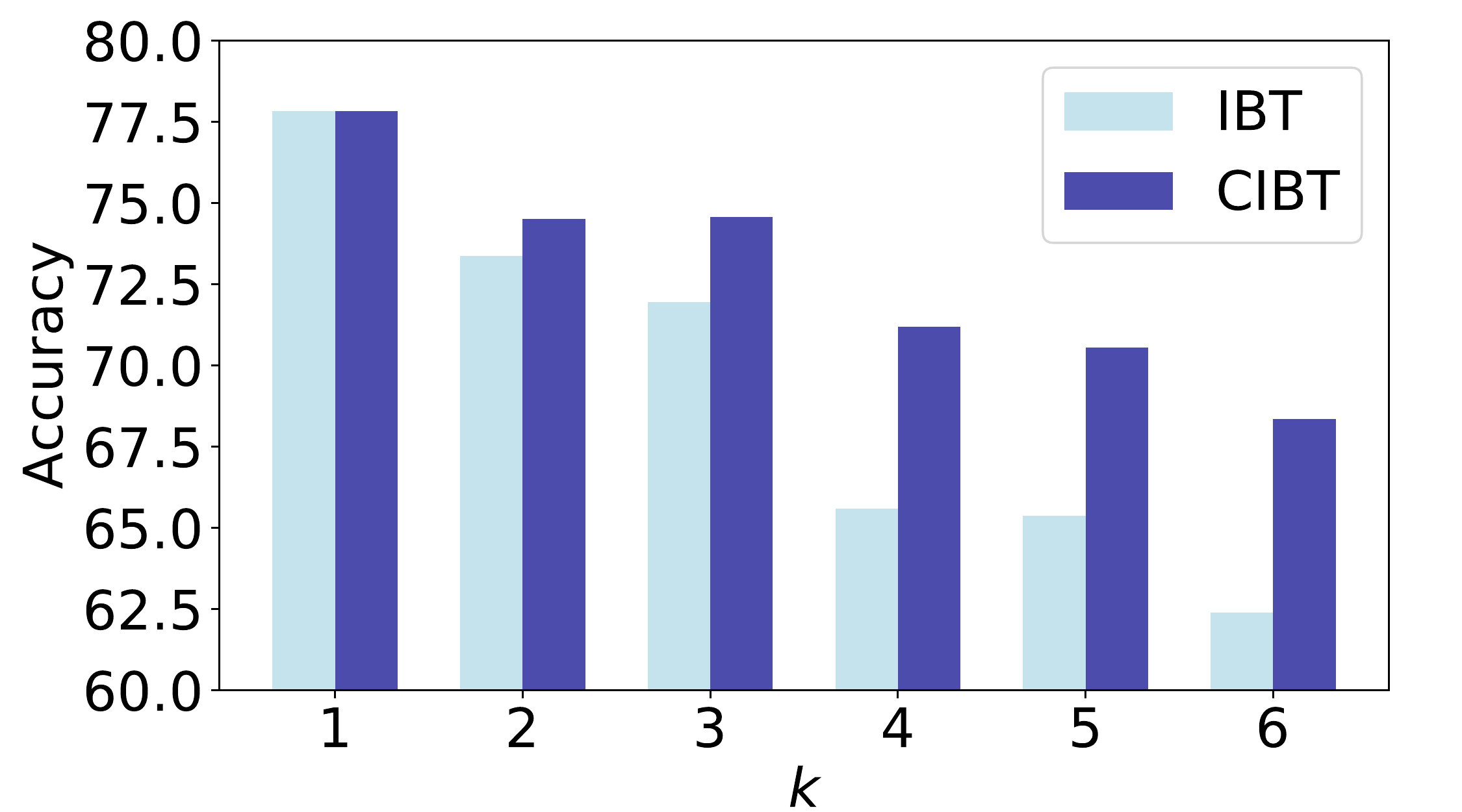}
\caption{Performance on different subsets. This figure indicates that curriculum learning is more beneficial to difficult data (larger $k$) than simple data (smaller $k$).}
\label{influence_of_each_curriculum}
\end{figure}

Table \ref{cl_results_summary} shows the main results. CIBT consistently outperforms IBT in all three data splits (MCD1/MCD2/MCD3) and all hyperparameter settings ($c$ =$2000 / 2500 / 3000 / 3500 / 4000$). For MCD1/MCD2/MCD3 tasks of CFQ, CIBT outperforms IBT by 1.8\%/11.3\%/5.8\% accuracy scores.
An interesting observation is that incorporating curriculum learning into iterative back-translation brings more stable performance:
IBT performs much worse on MCD2 (57.8\%) than on MCD1/MCD3 (64.8\%/64.6\%), while CIBT's improvement on MCD2 (11.3\%) is much better than on MCD1/MCD3 (1.8\%/5.8\%), making the performances on MCD1/MCD2/MCD3 comparable (66.6\%/69.1\%/70.4\%).

\subsubsection{Curriculum learning is more beneficial to difficult data than simple data.}
Figure \ref{influence_of_each_curriculum} further illustrates how curriculum learning improves the performance.
We use $k$ to denote test cases that would be put into the first $k$ curriculums, then obtain 6 subsets of test data: $T_1$, $T_2$, ..., $T_6$.
We have $T_1\subset T_2\subset ... \subset T_6$.
As we can observe in Figure \ref{influence_of_each_curriculum}, on the simplest subset ($T_1$), the performances of IBT and CIBT are very close.
If we consider $T_2$, a subset which is a bit more difficult than $T_1$, CIBT performs slightly better than IBT.
Similarly, we observe that: more difficult a subset of test data is, more gains CIBT brings.
Therefore, we can conclude that curriculum learning is more beneficial to difficult data than simple data.

\section{Related Work}

\subsection{Compositional Generalization}
\label{sec:related_cg}
Recently, exploring compositional generalization of DNN models has attracted a large attention on different topics, e.g., visual-question answering \cite{hudson2018compositional}, algebraic compositionality \cite{veldhoen2016diagnostic,saxton2018analysing}, and logic inference \cite{bowman2015tree,mul2019siamese}.
In this work, we focus on compositional generalization in language, which is benchmarked by SCAN \cite{lake2018generalization} and CFQ \cite{keysers2020measuring} tasks.

While mainstream DNN models are proved to exhibit limited compositional generalization ability in language \cite{lake2018generalization,hupkes2020compositionality,keysers2020measuring,furrer2020compositional}, there have been many efforts to address this limitation, including
design specialized architectures~\cite{russin2019compositional,li2019compositional,gordon2019permutation,guo2020hierarchical,liu2020compositional}, data augmentation~\cite{jia-liang-2016-data,andreas-2020-good}, meta-learning~\cite{lake2019compositional,nye2020learning} and pre-training models~\cite{furrer2020compositional}.
Comparing to these previous work, this paper incorporates semi-supervised learning into compositional generalization, which is task-agnostic and model-agnostic, thus improving the performance on both SCAN and CFQ benchmarks.

\subsection{Iterative Back-translation}
The idea of iterative back-translation dates back at least to \emph{back-translation}, which simply augments parallel training data using pseudo-parallel data generated from only target-side monolingual data.
Back-translation has been proven effective in statistical machine translation \cite{goutte2009learning,bojar2011improving} and neural machine translation \cite{sennrich2016improving,edunov2018understanding,imamura2018enhancement,xia2019generalized,wang2019improving}.
Iterative back-translation \cite{hoang2018iterative,cotterell2018explaining,lample2018unsupervised,conneau2019cross} is an extension of back-translation, in which forward and backward direction models generate pseudo-parallel data for each other and improve each other, bringing stronger empirical performance.

To our best knowledge, this paper is the first to study iterative back-translation from the perspective of compositional generalization.
We demonstrate that iterative back-translation can help seq2seq models make better generalizations to much more combinations, thus broadening the understanding of iterative back-translation.

\section{Conclusion}
In this paper, we revisit iterative back-translation from the perspective of compositional generalization.
We find that iterative back-translation is an effective method that exploits a large amount of monolingual data to enable seq2seq models better generalize to more combinations beyond limited parallel data.
We also find that iterative back-translation has the ability to increasingly correct errors in pseudo-parallel data, thus contributing to its success.
To encourage this behaviour, we propose curriculum iterative back-translation, which explicitly improves the quality of pseudo-parallel data, thus further improving the performance.


\bibliography{aaai21}
\end{document}